\documentclass[review]{elsarticle}

\usepackage{lineno,hyperref}
\usepackage{amsmath}
\usepackage{color}
\usepackage{xcolor}
\usepackage{colortbl}
\usepackage{multirow}
\usepackage{multicol}
\modulolinenumbers[5]

\journal{Journal of \LaTeX\ Templates}

\begin{document}

\begin{frontmatter}

\title{Learning with Rethinking: Recurrently Improving Convolutional Neural Networks through Feedback }


\author[address1]{Xin Li}
\ead{lixin08@ocrserv.ee.tsinghua.edu.cn}
\author[address2]{Zequn Jie}
\ead{jiezequn@u.nus.edu}
\author[address2]{Jiashi Feng}
\ead{elefjia@nus.edu.sg}
\author[address1]{Changsong Liu\corref{mycorrespondingauthor}}
\ead{lcs@ocrserv.ee.tsinghua.edu.cn}
\author[address2]{Shuicheng Yan}
\ead{eleyans@nus.edu.sg}
\cortext[mycorrespondingauthor]{Corresponding author}

\address[address1]{State Key Laboratory of Intelligent Technology and Systems, Tsinghua National Laboratory for Information Science and Technology, Tsinghua National Laboratory for Information Science and Technology, Department of Electronic Engineering, Tsinghua University, Beijing 100084, China}
\address[address2]{Department of Electrical and Computer Engineering, National University of Singapore, Singapore 117583, Singapore}

\begin{abstract}
Recent years have witnessed the great success of convolutional neural network (CNN) based models in the field of computer vision. CNN is able to learn hierarchically abstracted features from images in an end-to-end training manner. However, most of the existing CNN models only learn features through a feedforward structure and no feedback information from top to bottom layers is exploited to enable the networks to refine themselves. In this paper, we propose a ``Learning with Rethinking'' algorithm. By adding a feedback layer and producing the emphasis vector, the model is able to recurrently boost the performance based on previous prediction. Particularly, it can be employed to boost any pre-trained models. This algorithm is tested on four object classification benchmark datasets: CIFAR-100, CIFAR-10, MNIST-background-image and ILSVRC-2012 dataset. These results have demonstrated the advantage of training CNN models with the proposed feedback mechanism.
\end{abstract}


\end{frontmatter}


\section{Introduction}
Object recognition aims at automatically assigning labels of object categories from a finite label collection to a given image. It is a fundamental problem in the field of computer vision and also a core technique for  many  applications~\cite{girshick2014rich,long2015fully,sun2014deep}. Various algorithms for object recognition have been developed in the past decades which can be roughly summarized into the following standard pipeline: first a variety of handcrafted features are extracted; then the features are fed into some sophisticated feature encoding or transformation (\emph{e.g.} dimension reduction, feature pooling) procedures; finally those high-level features are classified with trained sophisticated classifiers. Though many works (\emph{e.g.} SIFT~\cite{ng2003sift}, LBP \cite{he1990texture}, HOG \cite{dalal2005histograms}, Gabor\cite{marvcelja1980mathematical})  focus on developing better handcrafted features, the feature is still undoubtedly the major bottleneck for improving the performance of object recognition.

In recent years, great progress has been achieved in object recognition which is arguably attributed to the availability of larger datasets for training more sophisticated models and greater computation resources, and more importantly the application of deep learning algorithms.

The convolution neural network (CNN) -- a popular example of deep learning algorithms -- adopts a deep architecture that consists of many stacked convolutional and fully-connected layers. Such an architecture  is specifically designed for solving computer vision related problems \cite{le1990handwritten,lecun1998gradient,ciresan2012multi,yin2013icdar, krizhevsky2012imagenet}  and  has also seen many other successful applications. The designed architecture of CNN is end-to-end trainable and  is able  to automatically learn features performing well for specific targets at different abstraction levels. With these high-level  features, it is possible to classify images accurately with a simple classifier. 
Nowadays, CNN-based algorithms have achieved the state-of-the-art results on many challenging tasks \cite{girshick2014rich,girshick2015fast,ren2015faster,taigman2014deepface,sun2014deep}. 

However, the simple feedforward architecture cannot well handle some challenging cases of object recognition. For instance, it has been observed that the powerful GoogLeNet~\cite{szegedy2015going} often fails in recognizing small objects in an image. 
Another challenging case for the feedforward deep architecture is the fine-grained object recognition~\cite{zhang2014part} where the differences among different fine-grained categories are quite subtle. Distinguishing fine-grained categories requires the CNN based models to extract features from the most discriminative regions. Therefore, part annotations are usually utilized to assist fine-grained image classification~\cite{xiao2015application}. Through empirical statistics on the classification errors, we find that the network is able to predict several candidate categories that include the correct one with high confidence. However, making the correct final decision on the single category  is difficult for the network based models, due to the distraction from other candidate categories.
Motivated by the above observations, we propose a novel ``Learning by Rethinking'' (LR) algorithm in this paper: instead of making the final decision based on one-pass of the data through the network, we introduce feedback connections and allow the network based models to ``re-think'' the decision and take the high-level feedback information into feature extraction. Benefiting from the feedback, the model is able to extract more discriminative low-level features with the guidance from the high-level information.

\begin{figure}[!htb]
	\hspace{-1.0cm}
	\includegraphics[width=135mm]{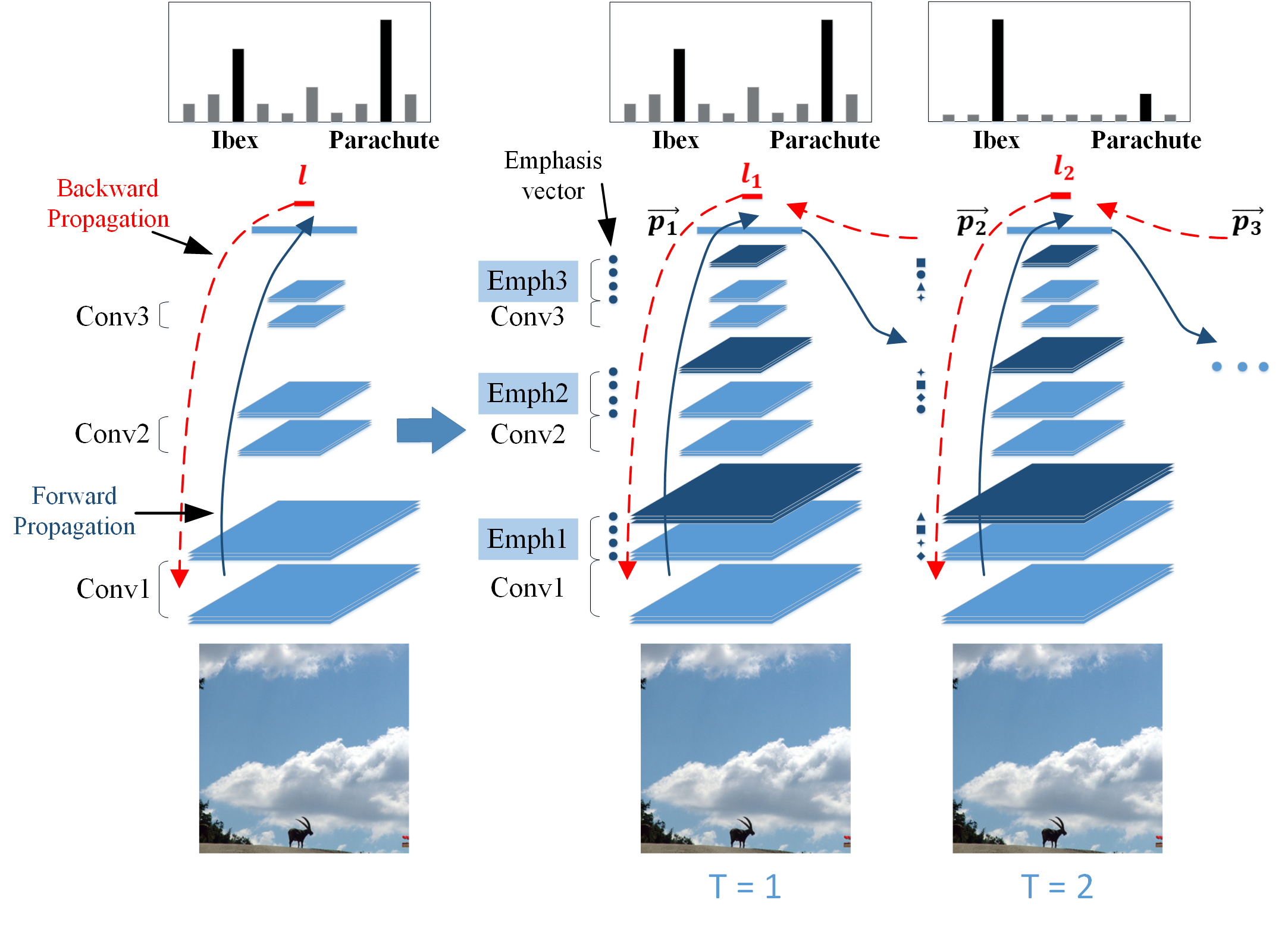}
	\caption{Illustration of the time-unfolded overall pipeline. With a pre-trained model, several emphasis layers are inserted. Initial emphasis vectors in $T=1$ are fixed for all $1$, and other emphasis vectors are calculated from feedback layers in the following iterations. The emphasis layers then alter the response of corresponding feature maps through the inserted emphasis layers. The total loss during training is the sum of all $l_t$ with equal weights. The blue solid and red dash arrows refer to the forward pass and the backward pass respectively. }
	\label{fig:pipeline}
\end{figure}

We propose two new types of layers -- the ``feedback'' layer and the ``emphasis'' layer -- to serve as the channel for transferring the feedback information. The feedback layer connects one top layer to one specific bottom layer in the network. The emphasis layer produces different weights based on such top-down information for the feature maps in the connected bottom layer. The proposed ``Learning with Rethinking'' algorithm exploits the fed back posterior probability of candidate object categories in the feedback layer, and endows the network model with the ability to ``re-think'' the decision during training. A new prediction will be made with consideration of previous prediction.

Figure \ref{fig:pipeline} provides the overall pipeline of the ``Learning with Rethinking'' algorithm in a time-unfolded manner for illustration purpose. Here we take the network-in-network (NIN) network as a basic CNN structure and illustrate how we build the proposed ``Learning with Rethinking'' network through augmenting the existing neural network architecture. The small ``ibex" in the image is misclassified as ``parachute" by a classic feedforward CNN (as shown in the left part where the probability of ``parachute" is larger than ``ibex"). In contrast, by further exploiting the posterior probabilities in the top layer before making final decision, the ``Learning with Rethinking'' algorithm recurrently adjusts the feature maps in hidden layers through feedback connection and identifies the correct category from other distracting categories. We will detailedly describe this pipeline in Section \ref{sec:learning_with_rethinking}.

The remaining content is organized as follows. In Section~\ref{sec:related_work}, we review some related works. Section~\ref{sec:learning_with_rethinking} describes the architecture and other details of our ``Learning with Rethinking" algorithm. Section~\ref{sec:experiments} presents the experimental results. Finally, Section~\ref{sec:conclusion} concludes the paper.

\section{Related Work}
\label{sec:related_work}

It has been twenty years since Lenet was first applied to OCR in 1990~\cite{le1990handwritten}. Many algorithms have been developed to improve the performance of CNN, although the basic framework of CNN has not changed much ever since it was proposed. 
The large object recognition data set ILSVRC2012, also known as ImageNet~\cite{deng2009imagenet}, has greatly propelled the progress in this area. Some most well-known progress in CNN structure has been made along with the continuous improvements of the performance on ImageNet data set. After AlexNet was proposed on ILSVRC2012, there are some remarkable advances in CNN architecture \cite{zeiler2014visualizing,DBLP:journals/corr/LinCY13,simonyan2014very,szegedy2015going,he2016deep}. And also, there are some task-specific modifications on CNN structure \cite{cirecsan2011committee,sermanet2011traffic,sun2014deep,barat2016string,cao2015large}. For example, in multi-resolution CNN~\cite{cirecsan2011committee,sermanet2011traffic,sun2014deep}, combining features in lower layers leads to a more detailed representation of an input image. MOP-CNN \cite{gong2014multi} is another algorithm proposed to extract more powerful features. With a combination of VLAD and CNN, MOP-CNN extracts a multi-scale and robust feature. This algorithm does not actually change CNN structure, but utilizes a pre-trained CNN model and modifies the feature extraction procedure.

Besides of exploring the overall structure of CNN, there are also many works that focus on each component of CNN. Locally connected layer~\cite{taigman2014deepface, yim2015rotating} loose the weight sharing constraint in normal convolution layer, and is suitable for face related tasks. 
Leaky ReLU~\cite{graham2014spatially} adds a negative slope to the normal ReLU, to preserve information discarded by ReLU. PReLU~\cite{he2015delving}   further enhances this by making the negative slope learnable.
Spatial Pyramid Pooling (SPP)~\cite{he2014spatial} extends max-pooling by enables CNN to avoid input warping or resizing and still produces fixed-length features. 
Inspired by Dropout~\cite{srivastava2014dropout}, DropConnect~\cite{wan2013regularization} regularize the CNN by randomly setting a subset of weights to zero within each layer. 
Spatial Dropout~\cite{tompson2015efficient} randomly sets some feature maps to zero entirely. DropSample~\cite{yang2016dropsample} randomly selects low confidence samples during training according to the output of CNN.
The commonly used fully-connected layer can be transformed into convolution layer with kernel size $1$, as shown in~\cite{DBLP:journals/corr/LinCY13}. With this transformation, CNN can take the input of any size and output classification maps.

Considerable works have been devoted to improving the performance of a CNN model through modifying its architecture. However, all these algorithms are still founded on a single feedforward pass of samples. Rare effort has been made to \emph{recurrently} improve the performance of a CNN model. In this work, we argue that a recurrent recognition processing is more consistent with the mechanism embedded in the human brain for visual processing, motivated by neural science research~\cite{buffalo2010backward,gilbert2007brain,hupe1998cortical,bastos2015dcm}.

Based on the analysis of response latencies to a newly-presented image, there are two stages of visual processing: a pre-attentive phase and an attentional phase, corresponding to feedforward and recurrent processing respectively~\cite{lamme2000distinct}. And the feedback connections play an important role in the attentional phase~\cite{gilbert2007brain,hupe1998cortical}. Different with feedforward connections which directly carry information, the feedback connections primarily play a modulatory role~\cite{bastos2015dcm}. Experiments have shown that recurrent processing contributes to making object recognition in degraded images more robust~\cite{wyatte2012limits}.

The idea of recursive or recurrent neural network has a long history, and recursive neural network (RNN) is successful in modeling temporal and sequential data~\cite{graves2009novel, donahue2015long}. Several works consider employing recursive neural network on processing a single image.
Eigen \emph{et al.}~\cite{eigen2014understanding} proposed a recursive convolutional network in image classification, finding that too large recursion depth may result in inferior performance due to over-fitting. 
Ming Liang~\cite{liang2015recurrent} enhanced the recursive layer by taking feed-forward inputs into all un-folded layers, the recurrent connections are spatial within the same recursive layer.
Kim \emph{et al.}~\cite{kim2016deeply} propose a deep recursive convolutional neural network for image super-resolution. The recursion depth is much more larger, and all predictions from the intermediate recursion is utilized to obtain the final output.

Our ``Learning with Rethinking" algorithm differs from above recursive neural networks in that we combines the posterior probabilities in the top layer into next recursion. The ``Learning with Rethinking'' algorithm recurrently adjusts the feature maps in hidden layers through feedback connection and identifies the correct category from other distracting categories

The idea of refining prediction is similar to cascading, which is a multistage ensemble learning algorithm. The subsequent stages focus on refining predictions of previous stages~\cite{sun2013deep,li2015convolutional,timofte2016seven,ren2015faster}. For instance, state-of-the-art object detection algorithms adopt a two-stage pipeline~\cite{ren2015faster}. The region proposal network proposes object candidates in the first stage, and the detection network focus on classifying proposals in the following stage. 
Sun \emph{et al.}~\cite{sun2013deep} proposed three-stage cascaded convolutional neural networks for facial point detection, where the subsequent stage focus on giving more accurate keypoints estimation.
Li \emph{et al.}~\cite{li2015convolutional,qin2016joint} proposed three-stage cascaded convolutional neural networks for face detection, where the first two stage quickly reject easy background regions, and the third stage  carefully evaluates a small number of challenging candidates.
Timofte \emph{et al.}  employed a four-stage cascaded models to gradually refine the contents in image super-resolution. They kept the same settings for all the stages but models are trained per stage.

Our ``Learning with Rethinking" algorithm differs from above cascading algorithms in that we recurrently refine the same model in all stages.
In contrast, cascading algorithms needs to train a model for each stage.

The most related work on utilizing the recurrent neural network for object recognition would be dasNet~\cite{stollenga2014deep}. It makes use of a reinforcement learning strategy to iteratively adjust some weights of feature maps. And final classification results are made after several iterations. Our ``Learning with Rethinking" algorithm differs from dasNet in three major aspects. Firstly, we use a neural network to feedback information into lower layers, which is relatively easy to calculate. Secondly, we only use the posterior probability of previous feedforward pass as the feedback information, which is much more timely and spatially efficient. Thirdly, our algorithm can be regarded as a new further training algorithm which is easy to be applied to any pre-trained models, and will further boost the performance. Comparatively, dasNet needs to train from random initialization.

\section{Learning with Rethinking}
\label{sec:learning_with_rethinking}
In this section, we briefly review the conventional architecture of convolutional neural networks (CNN). Then we elaborate  how to incorporate the feedback mechanism into the existing CNN architectures and propose the ``Learning with Rethinking'' algorithm to improve the performance of CNN for object recognition. The basic idea of ``Learning with Rethinking'' is intuitive: in addition to the feedforward connections in a neural network, several feedback connections directed from a top layer to a certain bottom layer are also established to provide top-down information for object recognition. With the higher-level information from the top layer, the bottom layers can stay being informed of those categories in the training data that are misclassified and those the layers need extra effort to distinguish. Such information is fed back through the ``emphasis layer'' and the ``feedback layer'' devised in this work.

\subsection{Conventional Convolutional Neural Networks}
In the conventional convolutional neural network (CNN) architecture, multiple layers of different types (\emph{e.g.}, convolutional layers and pooling layers) are connected in a simply feedforward manner and the information only moves in one direction.
In particular, each layer takes a collection of feature maps output by the previous layers as input, and produces a set of new feature maps via convolution or pooling operations. The new feature maps are then fed into the next layer directly. By stacking multiple convolutional layers interlaced with pooling layers, CNN can extract features at different abstraction levels with increasingly larger receptive fields. One advantage of employing such a feedforward mechanism in the CNN architecture is that the involved operations in producing the feature maps (such as convolution and pooling) are computationally efficient   without a directed cycle. And the algorithms of back-propagating errors from top layers to bottom ones can be applied straight-forwardly to efficiently optimize the parameters of the CNN. However, such a feedforward mechanism also has a limitation, since each layer only interacts with its neighboring layers and the important top-down information cross different layers is lost. 

In the following subsections, we introduce a new network architecture that also allows feedback connection among different layers. We elaborate how such an architecture with a feedback mechanism can learn more discriminative feature representation to better solve the object recognition problems -- especially for those involving recognizing objects of small size or from fine-grained categories with subtle differences. 

Throughout the paper, we use the following notations for the simplicity of explanation. The feature maps are represented by a tensor of dimension $C \times M \times M$, where $C$ is the number of feature maps, and $M \times M$ is the spatial dimension of each feature map. We use $f^{\ell}$ to denote the input feature maps for the $\ell$-th layer and $f^{\ell}_{i,p,q}$ to denote the value of the $i$-th feature map at the position $(p,q)$, in the $\ell$-th layer.

\subsection{Feedback Layer}
\label{subsec:feedback_layer}
We introduce the feedback mechanism to the conventional CNN architecture through a new feedback layer. The feedback layer connects two layers that may not be neighboring to each other in a top-down direction. When an input sample passes through all the layers, instead of immediately making a  prediction based on the predicted posterior probability of the sample belonging to a specific category, a feedback layer is deployed to propagate the predicted posterior probability to the bottom layers to update the network. Intuitively, when a sample has similar posterior probabilities for two different categories, it is not easy to be classified. Instead of outputting the final prediction directly, a wiser way is to guide the previous layers based on the current posterior probabilities of these confusable categories, such that the bottom layers can be strengthened or weakened to produce more discriminative features specifically for those categories difficult to distinguish.

\begin{figure}[!htb]
	\centering
	\includegraphics[width=40mm]{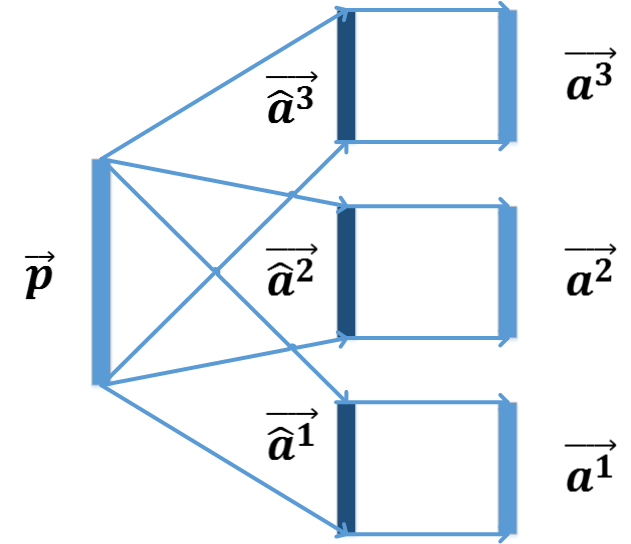}
	\caption{Illustration of the feedback layer. It takes posterior possibilities ${p}$ in the previous iteration as input, and produces the initial emphasis vectors 
with several fully connected layers. The initial emphasis vectors are then normalized to 
be emphasis vectors.}
	\label{fig:feedback_layer}
\end{figure}

Formally, suppose there are in total $N_{\text{class}}$ categories. Then for each sample, the network outputs $N_{\text{class}}$ posterior possibilities $\hat{p}_j,j=1,\ldots, N_{\text{class}}$, each of which denotes the possibility of the sample belonging to the corresponding category. The feedback layer is a fully connected layer whose parameters are denoted as $W^{s}$ and $b^{s}$ for producing the $s$-th emphasis vector. It takes posterior possibility $p$ as input and outputs $N_e$ emphasis vectors. The dimension of the $s$-th emphasis vector (here $s=1,\ldots,N_e$) is denoted as $C_s$, which equals the number of channels in the corresponding bottom layer. The $i$-th element in the $s$-th emphasis vector used to re-weight the $i$-th channel is computed as follows,
\begin{eqnarray}
\label{eqn:emphasis_grad}
\widehat{a}^{s}_i &=& \sum\limits_{j=1}^{N_{\text{class}}}W^{s}_{ij} \hat{p}_{j}+b^{s}_j, \\
\label{eqn:empahsis_normalize}
a^{s}_i &=&  \dfrac{C_s \cdot \exp(\widehat{a}^s_i)}{\sum_{j=1}^{C_s} \exp(\widehat{a}^{s}_j) }.
\end{eqnarray}
The initial emphasis vector $\widehat{a}^s$ computed in Equation \eqref{eqn:emphasis} is then normalized and weighted by the total channel number  $C^s$ in the emphasis vector in Equation \eqref{eqn:empahsis_normalize}. The emphasis vectors are then used to re-weight the feature maps in the layer connected to the feedback layer. Such normalization guarantees that coefficients in the emphasis vector have a mean value of $1$ such that the feature maps re-weighted by the emphasis vector have a magnitude at the same order as the feature maps without being augmented by the feedback and re-weighted. 

The computational cost in time and space in the feedback layer is negligible. For an output emphasis vector with a length of $C^s$, only $(C^s+1) \times N_{\text{class}}$ extra parameters are introduced. Each emphasis vector is able to adaptively rectify the feature maps -- through lifting contribution to certain layers and weakening the effects of other layers -- to produce more discriminative feature maps for the following object recognition. In the next subsection, we explain the role of the emphasis vectors in more details.

\begin{figure}[!tb]
	\centering
	\includegraphics[width=40mm]{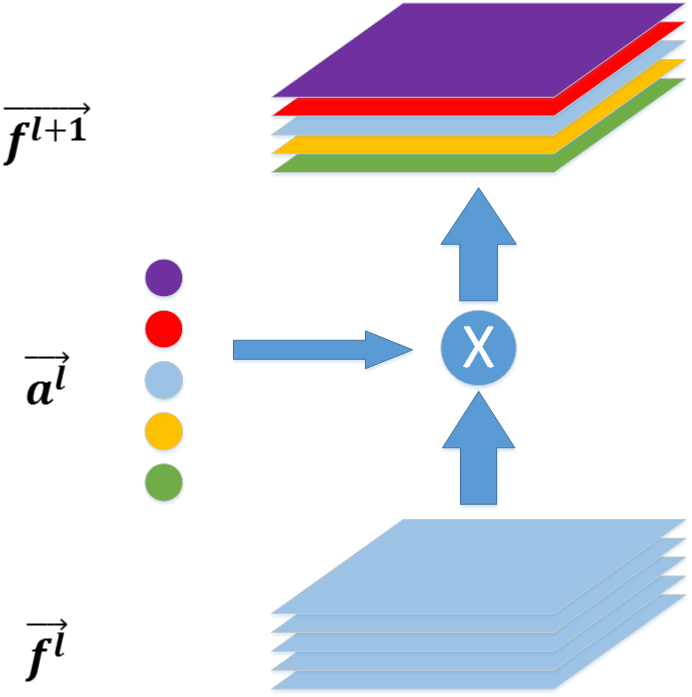}
	\caption{Illustration of the emphasis layer. The output feature maps $f^{\ell+1}$ are produced by re-weighting the input feature maps $f^{\ell}$ with the learned emphasis coefficients $a^{\ell}$. Different colors denote different values of the weights.}
	\label{fig:emphasis_layer}
\end{figure}

\subsection{Emphasis Layer} 
\label{subsec:emphasis_layer}
To adaptively re-weight different feature maps in a specific layer, an emphasis layer is introduced in our proposed Learning-with-Rethinking network. The emphasis layers take the emphasis vectors as well as the feature maps as inputs and outputs the re-weighted feature maps. More concretely, the $i$-th channel $f^{\ell}_{i}$ in the $\ell$-th layer is weighted by the corresponding emphasis coefficient $a_{i}$ by multiplying $a_{i}$ with $f^{\ell}_{i}$ : with $a_{i}>1$, the $i$-th channel is enhanced; and the channel is suppressed with $a_{i}<1$. All the emphasizing coefficients in the $\ell$-th layer form an emphasizing vector $a^{s}$. The intuition of this emphasis layer comes from the human visual mechanism. It is believed that the feedback connections primarily play a modulatory role. This structural augmentation enables the Learning-with-Rethinking network to selectively emphasize some discriminative features, and suppress the feature maps causing confusion in the recognition. The emphasis procedure can be formally written as

\begin{equation}
\label{eqn:emphasis}
 f^{\ell+1}_i = a^s_i f^{\ell}_i.
\end{equation}
Figure \ref{fig:emphasis_layer} illustrates such operation conducted by the emphasis layer.

\subsection{Architecture of the Learning with Rethinking Network}
\label{subsec:architecture}
With the feedback layer and the emphasis layer, we build the proposed Learning-with-Rethinking network through augmenting the existing neural network architecture. Here we take the network-in-network (NIN) network as a basic CNN structure and illustrate how we can build and train the corresponding Learning-with-Rethinking NIN network LR-NIN.  Figure \ref{fig:pipeline} provides the  overall pipeline of the LR-NIN in a time-unfolded manner for illustration purpose. The LR-NIN network is constructed as follows.
 
First, we pre-train the NIN model without feedback connection to obtain an initial model of LR-NIN. Then, we build three emphasis layers that are connected to three different convolution layers. Each emphasis layer takes its corresponding emphasis vector from the feedback layer as input (ref. Section~\ref{subsec:feedback_layer}), and produces the emphasis vectors to re-weight the produced feature maps of each convolution layer (ref. Section \ref{subsec:emphasis_layer}). Such information feedback in the LR-NIN is repeated for $T$ times in total to train the overall network. 
 
In our implementation, all the coefficients in the emphasis vector are initialized as $1$, and the emphasis layer does not change the feature maps at the initial stage. In the training phase, LR-NIN recursively feeds back the posterior possibility at the $(T-1)$-th step to guide the operation at the $T$-th step. 
 
It is obvious that increasing the number of recursive steps for information feedback allows the bottom layers to receive richer top-down information, but   the training time cost will also be increased accordingly. We observe from experiments that after $T>2$ the performance improvement is incremental. In order to trade off the time cost and the final performance, we empirically set $T=2$ in the training phase of LR-NIN. 
 
LR-NIN only introduces very few extra parameters compared with NIN: LR-NIN adds three emphasis layers after the three convolution layers whose kernel size is greater than $1$. In this case, only 58k extra parameters are introduced on CIFAR-100, amounting to only $4\%$ increase in the total  number of network parameters.

Similar to other recurrent neural networks, the LR-NIN network is trained by unfolding it into a very deep feedforward network, and is optimized by backpropagation through time (BPTT) algorithm \cite{rumelhart1985learning} with Stochastic Gradient Descent (SGD). Due to gradient vanishing probolem \cite{lee2015deeply,hochreitergradient,graves2012supervised}, the error signals propogated back tend to either blow up or vanish. This leads to failure in learning long time dependencies (training deep networks) sometimes. Inspired by the Deeply Supervised Network (DSN), we provide intermediate supervision for the intermediate predictions. As shown in Figure \ref{fig:pipeline}, there are $T$ cross-entropy losses corresponding to the $T$ feedforward passes. And the gradients of the losses are summed to be final gradients.

Formally, the loss $L$ to be optimized is a combination of the per-iteration cross-entropy loss $\ell_t$,
\begin{equation}
\label{eqn:loss}
 L = \sum\limits_{t=1}^{T}  \ell_{t} = \sum\limits_{t=1}^{T} (\sum\limits_{j=1}^{N_{class}} -p_j log\hat{p_j}).
\end{equation}

During feedforward propagation, the coefficients in the emphasis vectors are produced by the feedback layer, as explained in Section \ref{subsec:feedback_layer}. During back propagation, the gradients of the input feature maps $f^{\ell}$ and the emphasis vectors $a^s_i$ can be calculated via the chain rule:
	\begin{eqnarray}
	\dfrac{\partial \text{Loss}}{\partial f^{\ell}_i} &=& \dfrac{\partial \text{Loss}}{\partial f^{\ell+1}_{i}} a^{\ell}_{i}, \\
	\label{emphasis_grad1}
	\dfrac{\partial \text{Loss}}{\partial a^{\ell}_i} &=& \sum\limits_{pq} \dfrac{\partial \text{Loss}}{\partial f^{\ell+1}_i} f^{\ell_{ipq}}.
	\label{emphasis_grad2}
	\end{eqnarray}
The training and testing speed is roughly $T$ times slower than the original model. However, during training LR models, we can initialize the LR model with a well trained baseline model, and do not need to train the model from scratch. This makes the total training time reduced.

With these structural augmentations, LR enables the bottom layers in an existing model to be aware of the current classification prediction in the top layers. Then features are emphasized adaptively in the following iteration. In this way, the network is able to distinguish between confusing categories and yield better classification performance.

\section{Experiment}
\label{sec:experiments}
\subsection{Overall Settings}
We evaluate the performance of the LR algorithm on four benchmark datasets for image classification: CIFAR-100 \cite{krizhevsky2009learning}, CIFAR-10 \cite{krizhevsky2009learning}, MNIST-background-image \cite{larochelle2007empirical} and ILSVRC-2012 \cite{ILSVRC15}. Four pre-trained CNN models are employed as the baseline models which include NIN \cite{DBLP:journals/corr/LinCY13}, R-CNN \cite{liang2015recurrent}, LeNet \cite{lecun1998gradient}, VGG-Net~\cite{simonyan2014very}. We implement the LR algorithm on the Caffe platform \cite{jia2014caffe}.

Throughout the experiments, we fix the step of recursive feedback as $T=2$. On CIFAR-100, we also report the performance of the LR algorithm with $T>2$ in order to investigate the effect of $T$ on the final performance. Batch size is fixed as $128$ on CIFAR-10, CIFAR-100, MNIST-background-image and $32$ on ILSVRC-2012. 
The initial learning rate is set to $0.01$ on CIFAR100, CIFAR10, MNIST-background-image, and $0.00001$ on ILSVRC-2012. The momentum is set as $0.9$ in all the experiments. Weight decay of the L2 normalization is used as the regularization. 
The weight decay coefficient is set to $0.0001$ in all experiments. No weight decay is applied to any bias term. 
All these hyper parameters are not particularly tuned. And the dropout rates stay the same with the three publicly released models.

\subsection{CIFAR-100}
CIFAR-100 is a widely used benchmark dataset for image classification. There are in total $60{,}000$ color images of $100$ categories in the dataset. All the samples are split into $50{,}000$ for training  and $10{,}000$ for testing. The size of images in CIFAR-100 is  $32 \times 32$. NIN has been proven to be a successful CNN structure on CIFAR-100 \cite{DBLP:journals/corr/LinCY13,lee2015deeply}. We follow the same image pre-processing procedure used in NIN, \emph{i.e.} global contrast normalization and ZCA whitening.

We conduct three sets of experiments with different settings to evaluate the proposed LR algorithm. In the first experiment, we train a vanilla NIN model without data augmentation as the baseline. There are $3$ convolution layers with kernel size $5$,  each followed by $2$ convolution layers with kernel size $1$. Then we train an LR-CNN with the Learning-with-rethinking algorithm. The overall pipeline is the same as the one shown in Figure \ref{fig:pipeline}.  Three emphasis layers are added after each convolution layer with the kernel size of $5$. And corresponding feedback layers are added to produce emphasis vectors. 

In the second experiment, we train a new baseline model termed as LNIN. LNIN differs from NIN in the non-linear rectification unit. LNIN uses the leaky-ReLU to replace ReLU in NIN. In this setting, we train the LNIN model without data augmentation. It turns out that  Leaky-ReLU is a more effective non-linearity function on CIFAR-100 dataset than ReLU. We then train an LR-LNIN with the LR algorithm based on this LNIN baseline model, following the same procedure as in the first experiment.

In the third experiment, we use the pre-trained LNIN with data augmentation as the baseline model which  is named as LNIN-aug. Comparison with this baseline validates the effectiveness of our algorithm on further boosting models with even better performance. As for data augmentation, instead of using the heavy data augmentation used in sparse-cnn \cite{graham2014spatially}, we only use horizontal reflection. During training, we randomly flip the input image. In the test phase, the model makes predictions on both the original image and its mirror. Final classifications are given by simply averaging the two predicted posterior possibilities.

\begin{table}[!htb]
\centering
\caption{Comparison with baseline models on CIFAR-100. Our proposed ``Learning with Rethinking" algorithm clearly yields better performance for all the three settings.}
\renewcommand{\arraystretch}{1}
\begin{tabular}{l c c}
\hline
Model        & No. of Param. & Error (\%) \\
\hline
\multicolumn{3}{c}{\bfseries{ReLU, without data augmentation}} \\
\hline
NIN          & 0.98M        &    35.68 \\
ft-NIN          & 0.98M        & 35.11     \\
LR-NIN (T=2)  & 0.98M+0.058M     & \bf{33.32}     \\
\hline
\multicolumn{3}{c}{\bfseries{Leaky ReLU, without data augmentation}} \\
\hline
LNIN         & 0.98M         & 34.01\\
ft-LNIN          & 0.98M        &   33.30      \\
LR-LNIN (T=2)        & 0.98M+0.058M     & \bf{31.49}    \\
\hline
\multicolumn{3}{c}{\bfseries{Leaky ReLU, with data augmentation}} \\
\hline
LNIN-aug  	   &0.98M 			& 31.32 \\
ft-LNIN-aug          & 0.98M        & 30.14     \\
LR-LNIN-aug (T=2) & 0.98M+0.058M &  28.76 \\
LR-LNIN-aug (T=3) & 0.98M+0.058M &  \bf{28.36} \\
\hline
\end{tabular}
\label{table:cifar100_baseline}
\end{table}

\begin{figure*}[!htb]
	\hspace{-1.3cm}
	\includegraphics[width=145mm]{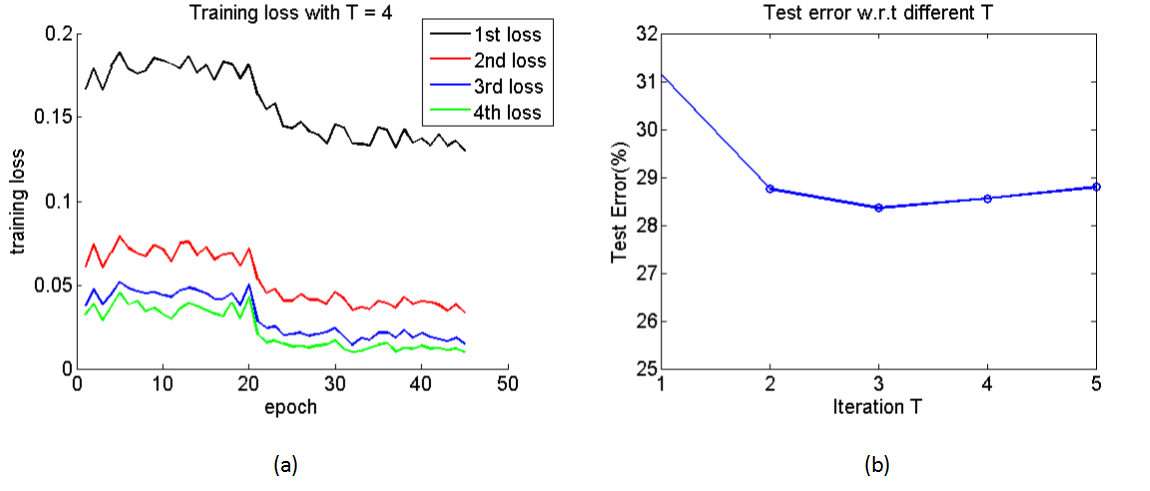}
	\caption{(a) The training loss with T=4. (b) Performance comparison with different T. The decrease of training loss in (a) shows our ``Learning with Rethinking'' algorithm works as expected. And the performance converges after T=3 in (b).}
	\label{fig:compareT}
\end{figure*}

Table \ref{table:cifar100_baseline} shows the comparison results of three experiments.
For a fair comparison, all baseline models are further trained for the same epochs as training the LR models, and corresponding models are referred with a prefix ``ft-''. We pre-trained each baseline model for $256$ epochs, and fine-tune with Learning-with-rethinking for another $256$ epochs. As shown in Table \ref{table:cifar100_baseline}, in all the three experiments, models trained with the LR algorithm not only outperform the pre-trained baseline model, but also outperform the further trained baseline models. Through comparing the baseline model and their corresponding further trained model, we can see that further training only brings minor improvements. But further training with LR could effectively improve the classification accuracy for near $2\%$.

We have also compared the performance of LR-LNIN-aug models with different $T$ values. As shown in Figure \ref{fig:compareT} (a), with a larger $T$ the training procedure still works as expected, and the training loss is decreasing with the same $T$. However, we can observe from Figure \ref{fig:compareT} (b) that the performance converges after $T=3$ on cifar-100 dataset, and a larger $T$ is not really necessary. Besides, a large $T$ leads to higher difficulty in training since we need $T$ times more computation than the baseline model.

\begin{table}[!htb]

\centering
\caption{Comparison with existing models with similar number of parameters and depth on CIFAR-100.}
\renewcommand{\arraystretch}{1}
\begin{tabular}{l c c c}
\hline
Model        & Input Size & No. of Param.  & Testing Error(\%) \\
\hline
\multicolumn{4}{c}{\bfseries{without data augmentation}} \\
\hline
Maxout \cite{goodfellow2013maxout}       & 32         & \textgreater5M & 38.57             \\
NIN \cite{DBLP:journals/corr/LinCY13}            & 32         & 0.98M               & 35.68             \\
DSN \cite{lee2015deeply}          & 32         & 0.98M                & 34.57            \\
RCNN-128 \cite{liang2015recurrent}    & 32         & 1.19M          &34.08*  \\
LNIN        & 32 & 0.98M         & 34.01\\
dasNet \cite{stollenga2014deep}       & 32         & \textgreater5M &  33.78            \\
\hline
LR-RCNN-128(T=2) & 32         & 1.19M+0.05M      &   31.95         \\
LR-LNIN(T=2) & 32         & 0.98M+0.058M      & \bf{31.49}             \\
\hline
\multicolumn{4}{c}{\bfseries{with data augmentation}} \\
\hline
NIN \cite{DBLP:journals/corr/LinCY13}           & 32         & 0.98M         &        33.53       \\
RCNN-128 \cite{liang2015recurrent}    & 32         & 1.19M          & 31.68*\\
LNIN        & 32         & 0.98M         &       31.32       \\
DeepCNet \cite{graham2014spatially}     & 96         & 25M           & 29.81             \\
DeepCNiN \cite{graham2014spatially}     & 96         & 34M           & 24.30        \\
\hline
LR-RCNN-128 \cite{liang2015recurrent}    & 32         & 1.19M +0.05M          & 30.72 \\
LR-LNIN-aug(T=3) & 32         & 0.98M+0.058M     & \bf{28.36}             \\
\hline
\end{tabular}
\label{table:cifar100_others}
\end{table}

We then compare our model with some state-of-the-art models of similar depth and model size. Table \ref{table:cifar100_others} shows the comparison results. 
Among the existing models in Table \ref{table:cifar100_baseline}, NIN~\cite{DBLP:journals/corr/LinCY13}, DSN~\cite{lee2015deeply} and LR-LNIN have comparable network depth and parameter number. RCNN~\cite{liang2015recurrent} models are much deeper. Maxout~\cite{goodfellow2013maxout} and dasNet~\cite{stollenga2014deep} employ more parameters. As for the DeepCNet and DeepCNiN~\cite{graham2014spatially}, input images are padded with zeros to  $96 \times 96$ ones, and the deep network models have $25$ million and $34$ million parameters respectively, which are much more than those in our model.

We also have evaluated our LR algorithm with RCNN-128 \cite{liang2015recurrent} as a baseline model. Four weight layers are inserted after every recurrent convolution layer. Because Liang~\cite{liang2015recurrent} do not release their Caffe implementation of RCNN-128 model on the CIFAR-100 dataset, we can not reproduce the reported accuracy and have to use our re-implementation model (marked with $\star$). The RCNN-128 model are further trained for the same epochs as training the LR-RCNN-128 to make a fair comparison.
As shown in Table \ref{table:cifar100_others}, our proposed model surpasses all the other models with moderate network depth and number of parameters.  
 Our LR-LNIN-aug model is only slightly inferior to DeepCNiN, which employs much more parameters. It is worth noting that our model beats DeepCNet when data augmentation is used -- our model only uses $1/25$ of the parameters of DeepCNet and only horizontal flip data augmentation.

\subsection{CIFAR-10}
CIFAR-10 is a dataset with the same image size and number of images as CIFAR-100. But its images are only from 10 categories. With fewer categories, the number of extra parameters introduced along with the feedback layer is only $1/10$ of the number in CIFAR-100. We evaluate the LR algorithm both with and without data augmentation. $T$ is fixed to 2.
All baseline models are further trained for the same epochs as training the LR models, and corresponding models are named with a prefix ``ft-''. As can be seen from Table~\ref{table:cifar10_baseline}, the experimental results show that models trained with the LR algorithm not only outperform the pre-trained baseline model, but also outperform the further trained baseline models.

\begin{table}[!tb]

\centering
\caption{Comparison with baseline models on CIFAR-10. The LR algorithm achieves better performance compared with well established baseline models.}
\renewcommand{\arraystretch}{1}
\begin{tabular}{l c c}
\hline
Model        & No. of Param. & Error(\%) \\
\hline
\multicolumn{3}{c}{\bfseries{Leaky ReLU, without data augmentation}} \\
\hline
LNIN           & 0.97M          &   9.92  \\
ft-LNIN 		& 0.97M 		& 9.74 \\
LR-LNIN(T=2)        & 0.97M+0.0058M     & \bf{9.13} \\
\hline
\multicolumn{3}{c}{\bfseries{Leaky ReLU, with data augmentation}} \\
\hline
LNIN-aug & 0.97M & 8.71	\\
ft-LNIN-aug &	0.97M & 8.56	\\
LR-LNIN-aug(T=2) & 0.97M+0.0058M &  \bf{7.67} \\
\hline
\end{tabular}
\label{table:cifar10_baseline}
\end{table}

\begin{table}[!htb]
	\centering
	
	\caption{Comparison with existing models on CIFAR-10. With fewer parameters, the proposed LR algorithm achieves comparable performance with state-of-the-art models.}
	\renewcommand{\arraystretch}{1}
	\begin{tabular}{l c c}
		\hline
		Model        & No. of Param.  & Testing Error(\%) \\
		\hline
		\multicolumn{3}{c}{\bfseries{without data augmentation}} \\
		\hline
		Maxout \cite{goodfellow2013maxout}        & \textgreater5M        &    11.68         \\
		NIN \cite{DBLP:journals/corr/LinCY13}     & 0.97M                      &   10.41         \\
		DSN \cite{lee2015deeply}                  & 0.97M                      &   9.69    \\
		DropConnect \cite{wan2013regularization}           & \textgreater5M        &    9.41           \\
		dasNet \cite{stollenga2014deep}           & \textgreater5M        &    9.22           \\
		RCNN-128 \cite{liang2015recurrent}       & 1.19M                      &   8.98      \\ \hline
		LR-LNIN(T=2)                             & 0.97M+0.0058M          &   9.13             \\
		LR-RCNN-128(T=2)                             & 1.19M+0.005M                     &   \bf{8.38}     \\
		\hline
		\multicolumn{3}{c}{\bfseries{with data augmentation}} \\
		\hline
		Maxout \cite{goodfellow2013maxout}             & \textgreater5M     &   9.38   \\
		NIN \cite{DBLP:journals/corr/LinCY13}          & 0.98M                    &  8.81  \\
		DSN \cite{lee2015deeply}                       & 0.98M                     & 7.97  \\
		RCNN-128 \cite{liang2015recurrent}            & 1.19M                      & 7.24  \\ \hline
		LR-LNIN-aug(T=2)                              & 0.98M+0.0058M              & 7.67  \\
		LR-RCNN-128-aug(T=2)                          & 1.19M+0.005M     & \bf{6.62}  \\
		\hline
	\end{tabular}
	\label{table:cifar10_others}
\end{table}

We compare our model with some state-of-the-art models on CIFAR-10 as shown in Table \ref{table:cifar10_others}. Our LR algorithm has improved the baseline RCNN-128 model with and without data augmentation.  And our LR-RCNN-128 model outperforms the state-of-the-art models with  a similar number of parameters and depth.

\subsection{MNIST-background-image}
MNIST-background-image is a variant of the popular MNIST digits dataset~\cite{larochelle2007empirical}. The gray digit image is surrounded with a gray image patch as background. It is more challenging than the original MNIST dataset. There are $12{,}000$ training images and $50{,}000$ testing images. The baseline LeNet model consists of two convolutional layers and one fully connected layer. The detailed structure of the LeNet model is shown in Table \ref{table:lenet}. Then we train an LR-LeNet with ``Learning with Rethinking" algorithm. Two emphasis layers are added after each convolution layer and corresponding feedback layers are added to produce emphasis vectors. 
We train the baseline model for $64$ epochs, and then train an LR-LeNet model for another $16$ epochs. Training the baseline LeNet model for another $16$ epochs do not give a better performance. 
The comparison results with other algorithms are shown in Table \ref{table:mnist_bakimg}. The baseline LeNet model outperforms other algorithms with a large margin, and our LR-LeNet model reduces the error rate by nearly $50\%$. To our best knowledge, LR-LeNet achieves the highest accuracy among all the reported results on MNIST-background-image dataset.

\begin{table}[!htb]
	\centering
	\caption{Structure of the baseline LeNet model.}
	\renewcommand{\arraystretch}{1}
	\begin{tabular}{l c c c}
		\hline
		name        & output size & channels & kernel size/stride \\
		\hline
		convolution1   & 24x24  &  20   &  5x5/1  \\
		max pool1      & 12x12  &  -   & 2x2/2  \\
		convolution2    & 8x8  &  50   &  3x5/1  \\
		max pool2       & 4x4  &  -   & 2x2/2  \\
		fully connected1 & -   &  500  & - \\
		fully connected2 & -   &  10  & - \\
		\hline \\
	\end{tabular}
	\label{table:lenet}
\end{table}

\begin{table}[!htb]
	\centering
	\caption{Comparison with existing models on MNIST-background-image. The LR-LeNet model achieves the state-of-the-art performance.}
	\renewcommand{\arraystretch}{1}
	\begin{tabular}{l c}
		\hline
		Model        & Error(\%) \\
		\hline
		DBN-3 \cite{larochelle2007empirical} 	      &   16.31  \\
		RBM \cite{wang2014attentional}           &   15.42  \\
		aNN-$\theta_{image}$  \cite{wang2014attentional} & 15.33  \\
		sDBN \cite{wang2014attentional}          &   14.34  \\
		PGBM \cite{wang2014attentional}          &   12.25  \\
		LeNet          & 6.92   \\
		\bfseries{LR-LeNet(T=2)}   & \bfseries{3.57}   \\
		\hline \\
	\end{tabular}
	\label{table:mnist_bakimg}
\end{table}

\begin{figure*}[!htb]
	\centering
	\includegraphics[width=80mm]{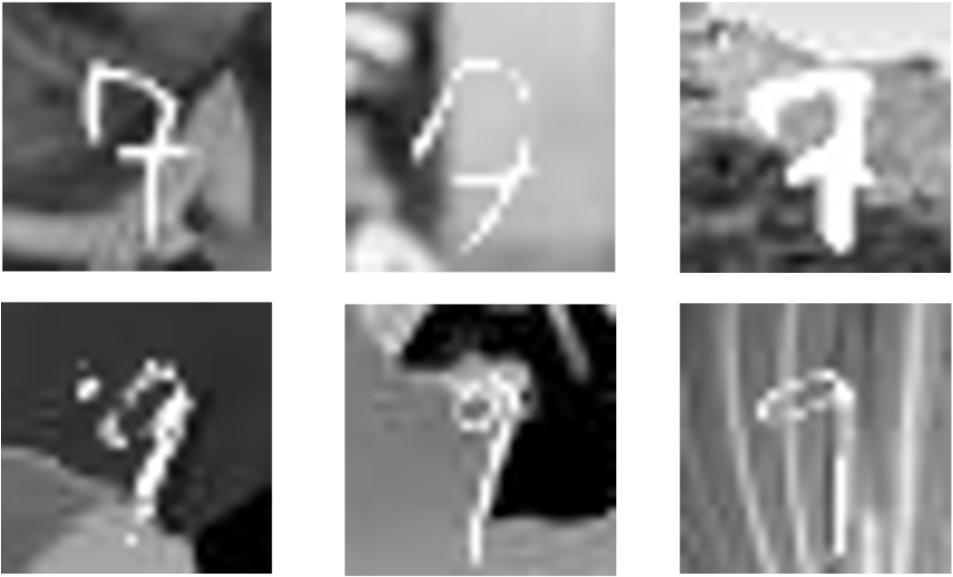}
	\caption{Some difficult examples of ``7"(top row) and ``9"(bottom row).}
	\label{fig:mnist_bakimg_samples}
\end{figure*}

\begin{figure*}[!htb]
	\hspace{-0.9cm}
	\includegraphics[width=135mm]{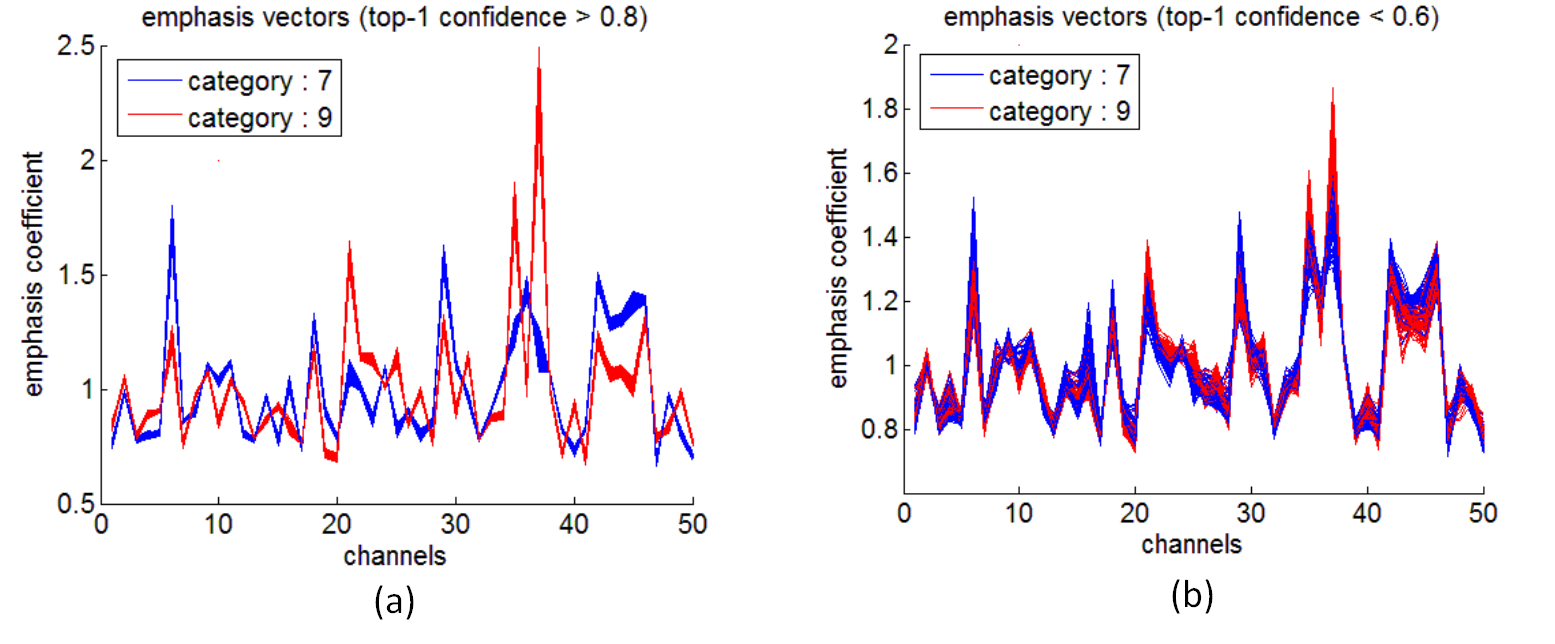}
	\caption{Emphasis vectors of samples from category ``7" and ``9". \textbf{(a):} Emphasis vectors of smaples with top1-confidence above 0.8. \textbf{(b):} Emphasis vectors of smaples with top1-confidence below 0.6.}
	\label{fig:minist_bakimg_emphasis_vector}
\end{figure*}

\begin{figure*}[!htb]
	\centering
	\includegraphics[width=120mm]{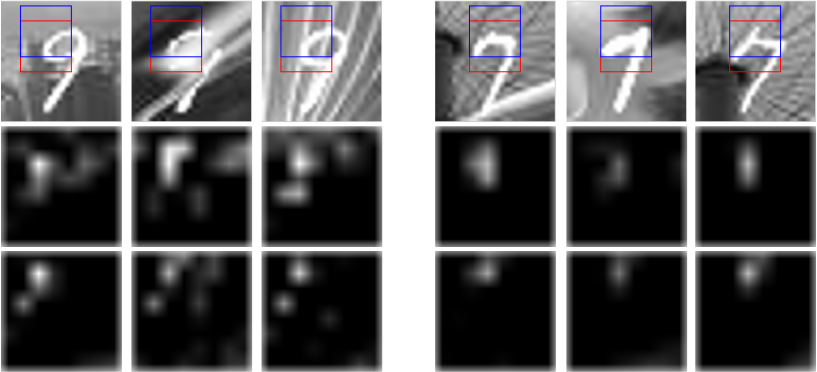}
	\caption{\textbf{Left:} The images and feature maps of category 9. \textbf{Right:} The images and feature maps of category 7. \textbf{Top Row:} Original input image. Red rectangles correspond to the receptive field of the maximum value in the $37$th feature map. Blue rectangles correspond to the receptive field of the maximum value in the $47$th feature map. \textbf{Middle Row:} Response of the $37$th channel. The responses magnitude of the $37$th channel in category 7 is usually weaker than in 9. \textbf{Bottom Row:} Response of the $47$th channel.}
	\label{fig:fm2check}
\end{figure*}

Benefiting from the small number of feature channels and categories, we are able to make a qualitative analysis of LR algorithm. We conduct visualization analysis on samples in category 7 and 9, which are commonly misclassified to each other. Several typical confusing images are shown in the left part of Figure~\ref{fig:mnist_bakimg_samples}. 
In Figure \ref{fig:minist_bakimg_emphasis_vector}, we visualize their emphasis vectors in the second emphasis layer. 
As shown in Figure \ref{fig:minist_bakimg_emphasis_vector} (a), when the confidence of top-1 candidate is higher, there are clear patterns of emphasis vectors. When the top-1 confidence is lower, \textit{i.e.} the predicted confidences of category 7 and 9 are at the same level, the emphasis vectors of both category 7 and 9 are similar, as shown in Figure \ref{fig:minist_bakimg_emphasis_vector} (b).
Emphasis vectors mainly enhance those features that increase the prediction confidence of the correct category. Therefore, in Figure \ref{fig:minist_bakimg_emphasis_vector} (a), the enhanced features differ for different categories. When the predicted confidences of category 7 and 9 are at the same level as shown in Figure \ref{fig:minist_bakimg_emphasis_vector} (b), emphasis vectors focus on enhancing the features which are the most beneficial in distinguishing these two categories, \textit{i.e.} the $37$th channel, and suppressing the features with weaker discriminability, \textit{i.e.} the $47$th channel.
We further visualize the corresponding feature maps in the middle row of Figure \ref{fig:fm2check}. The receptive field of the maximum value in the $37$th feature map is marked with red rectangles in input images. It seems that the $37$th feature map mostly responds to the left part of a ``blob". We also visualize the feature maps of the $47$th channel in the bottom row of Figure \ref{fig:fm2check}, which shows this channel responds to the top part of a ``blob", or rather, a curved horizontal line.

\subsection{ILSVRC 2012} 
The ILSVRC 2012 dataset is a much larger one than CIFAR-100, CIFAR-10 and MNIST-background-image. There are over 1.2 million color images in the training set, and 50k color images in the validation set. Top-5 accuracy on validation set is used as an evaluation metric. The VGG-Net is one of the top performed models on this dataset. These pre-trained VGG models are the \emph{de facto} basic component in many papers. 

There are two VGG-Net models released -- one has 16 layers and the other has 19 layers, termed here by VGG16 and VGG19 respectively. Both of them are pre-trained on the training data of ILSVRC 2012 with data augmentation. We use VGG16 as our baseline model due to its less training time cost but similar performance as VGG19 on ILSVRC 2012. An emphasis layer is added after each of the $13$ convolution layers. Also, $13$ corresponding feedback layers are added into the VGG16 network. This leads to $4.2M$ extra parameters, amounting to only $3\%$ of the total number of parameters. We trained the LR-VGG16 model for $10$ epochs. VGG16 is a well-trained model, and further training the baseline VGG16 model barely make any improvement.

Detailed comparisons are shown in Table \ref{table:imagenet}. This result on ImageNet validates the effectiveness of the LR algorithm on further boosting state-of-the-art models and shows its potential in large scale object classification tasks.

\begin{table}[!htb]

\centering
\caption{Comparison with the baseline models on ILSVRC 2012.}
\renewcommand{\arraystretch}{1}
\begin{tabular}{p{3cm}<{\centering}p{1.7cm}<{\centering}p{1.7cm}<{\centering}}
\hline
\multirow{2}{*}{Models}      & \multicolumn{2}{c}{Validation error(\%)} \\
                             & Single Crop & Multi Crop    \\
\hline
VGG16   &  10.05 & 8.8\cite{simonyan2014very}                    \\
LR-VGG16   &  9.16 & 8.02                   \\
\hline
\end{tabular}
\label{table:imagenet}
\end{table}

\begin{figure}[!htb]
	\hspace{-0cm}
	\includegraphics[width=120mm]{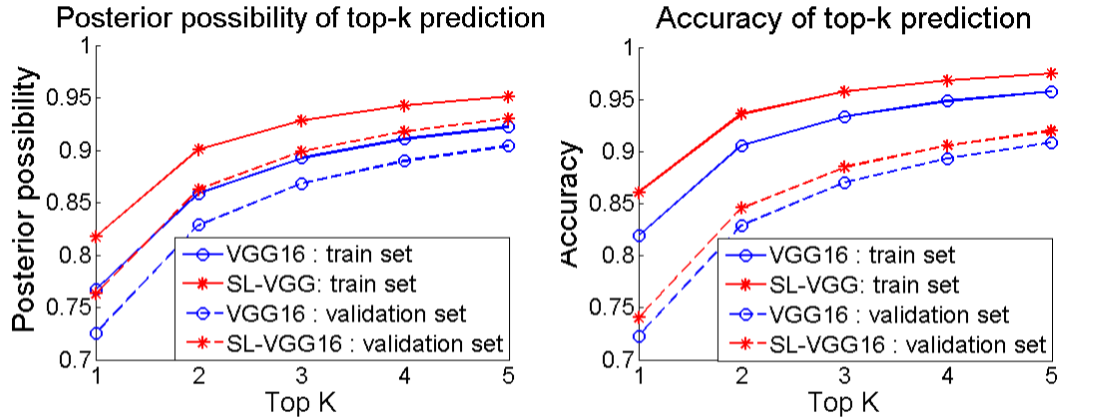}
	\caption{Posterior possibility (left) and accuracy (right) of top-k predictions on both training set and validation set in the ILSVRC dataset. 
		}
		\label{fig:imagenet}
\end{figure}

\begin{figure*}
	\centering
	\includegraphics[width=110mm]{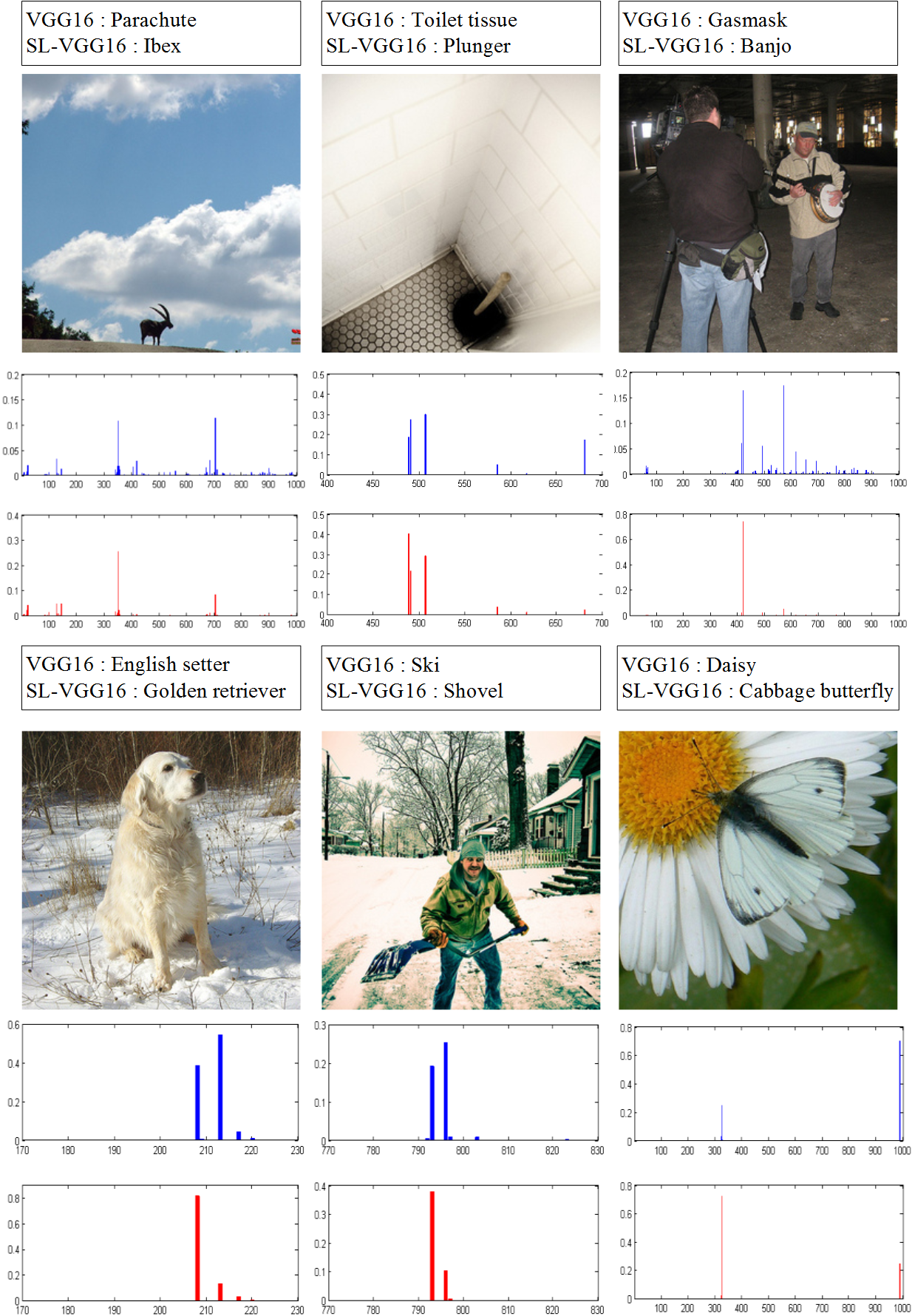}
	\caption{Illustration of corrected samples in the validation set. \textbf{Top row images}: Images are misclassified because objects are too small. \textbf{Bottom row images}: Images are misclassified to a similar category. \textbf{Blue histogram}: Posterior possibility produced by VGG16. Though images are misclassified, the network is able to predict several possible correct categories with high confidence.  \textbf{Red histogram}: Posterior possibilities produced by LR-VGG16. ``Learning with Rethinking" algorithm is able to choose the single correct category from the distracting candidate categories.}
		\label{fig:demos}
\end{figure*}

We here provide some analyses on the classification results of original VGG16 and LR-VGG16. As shown in Figure \ref{fig:imagenet}, the average posterior possibilities of top-1 prediction increase by $4 \sim 5\%$ on both training set and validation set. The improvement of top-k posterior possibilities demonstrates that LR-CNN is more ``certain" of its prediction. This shows that our algorithm can boost the model to make it fit   the training data better, and thus learn more information from training samples. By distinguishing  confusing categories,  the LR algorithm can improve the performance. The analysis of top-k accuracy in Figure \ref{fig:imagenet} also supports this observation. The top-1 accuracy of LR-VGG16 has surpassed the VGG16 model by more than $4\%$ on the training set, and a consistent improvement of $1\%$ is shown on the validation set. These statistical observations validate the effectiveness of ``Learning with Rethinking" on further boosting state-of-the-art models. Some examples of image classification results are shown in Figure \ref{fig:demos} with comparison between VGG16 and LR-VGG16. The results show that LR is more ``certain" of its prediction (\emph{i.e.} the entropy of the finally predicted probabilities is much smaller).

\section{Conclusion}
\label{sec:conclusion}
In this paper, we propose a ``Learning with Rethinking" algorithm for image recognition. The Learning with Rethinking algorithm feeds back posterior probability information from top layers to guide the bottom layers in their feature learning. Experiments on four benchmark datasets show that the Learning with Rethinking algorithm is able to further boost the well-established models with only a few parameters introduced. Particularly, experiments on benchmark datasets MNIST-background-image and ImageNet clearly demonstrate the advantage of the Learning with Rethinking algorithm in recognizing objects or categories with large inter-class similarity. Besides, our work also demonstrates that recurrently improving performance with feedback information is a promising direction. 

\section{Acknowledgment}
This work was supported by the National Basic Research Program of China (973 program) under Grant No. 2013CB329403 and the National Natural Science Foundation of China under Grant No.61471214.

\section{References}

\bibliography{lixin_bib}

\end{document}